\documentclass[conference]{IEEEtran}

\usepackage[hidelinks]{hyperref}
\hypersetup{
  pdfauthor={},
  pdftitle={Understanding Adaptive Normalization in Transformers through Differentiable Gating},
  pdfsubject={},
  pdfkeywords={Dynamic normalization, Transformers, Gumbel-Softmax, differentiable gating}
}
\usepackage{graphicx}
\usepackage{amssymb}
\usepackage{amsmath}
\usepackage{cite}        
\usepackage{booktabs}    

\begin{document}

\title{AutoNorm: Understanding Adaptive Normalization in Transformers\\
      through Differentiable Gating}
\author{
\begin{tabular}{cc}
Piyush Kaushik Bhattacharyya & Divyanshu Rai \\
\texttt{piyushbhattacharyya@gmail.com} &
\texttt{raidivyanshu3232@gmail.com} \\[0.5em]

Swastik Singh & Kumar Aakash \\
\texttt{singhswastik1103@gmail.com} &
\texttt{akashkumar221ank@gmail.com} \\[0.5em]

Ayush Ranjan & Krutika Verma \\
\texttt{ranjanayush881@gmail.com} &
\texttt{krutika.vermafcs@kiit.ac.in}
\end{tabular}
}
\maketitle
\begin{abstract}
Normalization is a critical component for stabilizing Transformer training, yet the choice between static strategies such as Layer Normalization (LN) and adaptive alternatives remains largely task-dependent. In this paper, we investigate a key optimization challenge in differentiable normalization gating. Our experiments show that, on relatively stationary vision tasks, the high gradient variance introduced by Gumbel--Softmax gating can hinder convergence of the routing mechanism, causing learned gates to underperform simple random selection. In contrast, on non-stationary language modeling and classification tasks, sustained gating diversity enables the model to learn more effective layer-wise normalization policies. Motivated by these observations, we propose \textbf{AutoNorm-S (Stabilized)}, a training strategy that mitigates optimization instability through a gate-freezing schedule. AutoNorm-S achieves competitive or improved performance across multiple benchmarks, outperforming adaptive normalization baselines on NLP datasets, including PTB and SST-2, while remaining competitive on standard vision benchmarks. These results suggest that decoupling normalization selection from optimization noise provides a practical and principled approach for adaptive normalization in Transformer architectures.
\end{abstract}


\section{Introduction}

Normalization techniques have become fundamental to the stability
and convergence of modern deep learning architectures. By normalizing
feature statistics within each layer, techniques such as Batch
Normalization (BN)~\cite{ioffe2015batch} and Layer Normalization
(LN)~\cite{ba2016layernorm} mitigate internal covariate shift and
stabilize gradient flow. In the era of Transformers~\cite{vaswani2017attention},
Layer Normalization has emerged as the de facto standard for stable
self-attention training across large model depths.

However, conventional normalization layers are fundamentally static,
applying identical transformations regardless of task context or input
distribution. This one-size-fits-all approach is increasingly
suboptimal as models encounter severe distribution shifts and diverse
data modalities. Prior works exploring adaptive normalization---such
as FiLM~\cite{perez2018film}, Switchable
Normalization~\cite{luo2019switchable}, and conditional
Meta-Normalization---typically rely on external metadata conditioning
or expensive architectural searches, lacking a self-sufficient
mechanism for on-the-fly adaptation. Recently, Zhu et
al.~\cite{zhu2025transformers} demonstrated that replacing
normalization entirely with a learnable tanh-based transformation
(Dynamic Tanh; DyT) can match or exceed LN in standard Transformers,
raising a natural question: \textit{can a model learn, at training
time, which normalization strategy is optimal per layer and per task?}

When we initially attempted to answer this using differentiable
Gumbel-Softmax gating---a framework we call \textbf{AutoNorm}---we
encountered a surprising and counter-intuitive phenomenon: while the
learned mechanism worked effectively for linguistic sequences, it
consistently failed on vision benchmarks, where a simple \textit{Random
Selector} often outperformed the learned policy. This unexpected
failure highlighted a deeper issue in how architectural selection
interacts with optimization dynamics.

In this work, we investigate this failure and identify it as a
\textbf{stabilization bottleneck} specific to stationary input
distributions. High gradient variance in the early stages of
differentiable gating prevents the selector from specializing before
the backbone features have matured. We introduce \textbf{AutoNorm-S
(Stabilized)}, which resolves this through a gate-freezing schedule
(Section~\ref{sec:freezing}), and demonstrate that the resulting
mechanism not only restores competitive performance in vision but
provides a significant inductive bias for NLP. Our contributions
progress from a specific architectural failure mode to a general
principle governing when and how adaptive normalization should be
deployed in Transformers.

For a given input tensor $X$, AutoNorm computes:
\begin{equation}
  \text{AutoNorm}(X) = w_{\text{DyT}}(X) \cdot \text{DyT}(X)
                     + w_{\text{LN}}(X)  \cdot \text{LN}(X)
  \label{eq:autonorm}
\end{equation}
where $w_{\text{DyT}}(X)$ and $w_{\text{LN}}(X)$ are adaptive weights
predicted by the NormSelector (Section~\ref{sec:normselector}),
satisfying $w_{\text{DyT}}(X) + w_{\text{LN}}(X) = 1$. These weights
are obtained through a lightweight gating network that uses a
differentiable Gumbel--Softmax
reparameterization~\cite{jang2017gumbelsoftmax}, allowing
gradient-based optimization of the discrete normalization selection.

\section{Literature Review}

\subsection{Normalization Techniques in Deep Learning}

Normalization is fundamental to training deep neural networks,
mitigating vanishing and exploding gradients and internal covariate
shift~\cite{ioffe2015batch,ba2016layernorm,ulyanov2016instance,wu2018group}.

\subsubsection{Batch Normalization (BN)}
Introduced by Ioffe and Szegedy~\cite{ioffe2015batch}, BN normalizes
activations across the mini-batch dimension and applies learnable scale
and offset. BN enables higher learning rates and acts as a regularizer
in CNNs but degrades with small batch sizes~\cite{wu2018group} and is
unsuitable for variable-length sequences.

\subsubsection{Layer Normalization (LN)}
Ba et al.~\cite{ba2016layernorm} proposed LN, which normalizes across
the feature dimension within each sample, independent of batch size.
LN is highly effective in Transformers but may underperform BN in
standard CNNs.

\subsubsection{Instance and Group Normalization}
Instance Normalization~\cite{ulyanov2016instance} normalizes per
channel over spatial dimensions, excelling at style transfer. Group
Normalization~\cite{wu2018group} divides channels into groups,
maintaining stable performance across batch sizes at the cost of
careful group-count selection.

\subsubsection{Adaptive Batch Normalization (AdaBN)}
Li et al.~\cite{li2018adaptive} proposed AdaBN for domain adaptation
by modulating batch-normalized statistics between source and target
domains---an early demonstration of adaptive normalization under
non-stationary distributions. AutoNorm extends this idea to
differentiable gated selection in Transformers.

\subsubsection{DeepNorm}
Wang et al.~\cite{wang2022deepnorm} introduced DeepNorm, enabling
Transformers to scale to 1,000 layers by stabilizing residual
variance. AutoNorm complements such scaling methods by providing
feature-dependent adaptability rather than static stability alone.

\subsubsection{Dynamic Tanh (DyT)}
Zhu et al.~\cite{zhu2025transformers} proposed replacing normalization
entirely with:
\begin{equation}
  \text{DyT}(x) = \boldsymbol{\gamma} \odot \tanh(\alpha x)
                + \boldsymbol{\beta}
  \label{eq:dyt}
\end{equation}
where $\alpha \in \mathbb{R}$ is a learnable scalar controlling input
saturation, and $\boldsymbol{\gamma}, \boldsymbol{\beta} \in
\mathbb{R}^{C}$ are learnable channel-wise scale and shift parameters.
DyT matches or exceeds LN on several Transformer benchmarks without
computing feature statistics at runtime. AutoNorm treats DyT and LN as
complementary candidates in a differentiable mixture, rather than
substitutes.

\subsubsection{Switchable Normalization}
Luo et al.~\cite{luo2019switchable} proposed Switchable Normalization
(SwitchNorm), which learns differentiable importance weights over BN,
LN, and IN. While SwitchNorm blends statistics computed simultaneously
from all three methods, AutoNorm uses Gumbel-Softmax gating to route
between fully computed normalization outputs and, critically,
\textit{identifies the optimization dynamics} that determine whether
learned switching is beneficial---a mechanism not studied by
SwitchNorm.

AutoNorm shares conceptual roots with FiLM~\cite{perez2018film} and
other conditional normalization methods. However, its Gumbel-Softmax
formulation enables fully differentiable autonomous selection between LN
and DyT without external metadata or task-specific conditioning.

\subsection{Dynamic and Adaptive Mechanisms in Neural Networks}

Adaptive mechanisms increase network flexibility during training and
inference~\cite{he2015prelu,ramachandran2017swish,ma2020metaacon,hu2018squeeze,chen2020dynamic}.
Key examples include adaptive activation functions---PReLU~\cite{he2015prelu},
Swish~\cite{ramachandran2017swish}, Meta-Acon~\cite{ma2020metaacon}---and
per-parameter adaptive optimizers: AdaGrad~\cite{duchi2011adagrad},
Adam~\cite{kingma2014adam}, AdaBelief~\cite{zhuang2020adabelief}.

\subsection{Transformer Architectures and Normalization Strategies}

Transformers~\cite{vaswani2017attention} rely on self-attention with
normalization, primarily LN. Pre-LN and Post-LN variants affect
training stability and scalability~\cite{xiong2020layernorm,wang2022deepnorm,bachlechner2020rezero}.
Alternatives include RMSNorm~\cite{zhang2019root} and
DeepNorm~\cite{wang2022deepnorm}. The interaction between normalization
strategy and optimization dynamics---the central focus of this
work---has received limited systematic study.

\section{Methodology}

\subsection{Differentiable Gating and Stabilization}
\label{sec:bottleneck_def}

We define input distribution non-stationarity in terms of the variability
of feature statistics across training batches. Specifically, datasets
exhibiting high variance in activation statistics or slow entropy
collapse of the gating distribution are considered non-stationary,
whereas datasets with stable feature distributions and rapid entropy
collapse (e.g., MNIST) are considered stationary. This definitional
grounding is essential for the design of the gate-freezing mechanism.

\begin{table*}[t]
\centering
\caption{Performance Comparison Across Vision Classification Datasets.}
\label{tab:vision_results}
\resizebox{\textwidth}{!}{%
\begin{tabular}{lccccccc}
\toprule
\textbf{Method}
  & \textbf{Val Acc $\pm$ std}
  & \textbf{Precision}
  & \textbf{Recall}
  & \textbf{F1-Score}
  & \textbf{Latency (s)}
  & \textbf{FLOPs}
  & \textbf{Rot.\ Acc$^\dagger$} \\
\midrule
\multicolumn{8}{c}{\textbf{MNIST}} \\
\midrule
AutoNorm-S          & \textbf{99.18 $\pm$ 0.10} & \textbf{99.20} & \textbf{99.18} & \textbf{99.19} & 0.0053 & 79{,}391{,}488 & \textbf{10.10} \\
FrozenLN            & 98.82 $\pm$ 0.13 & 98.80 & 98.81 & 98.80 & 0.0055 & 79{,}391{,}488 & 8.13 \\
AdaNorm             & 98.95 $\pm$ 0.12 & 98.92 & 98.94 & 98.93 & 0.0054 & 79{,}391{,}488 & 9.21 \\
Teacher (MLP)       & 97.98 $\pm$ 0.22 & 97.90 & 97.95 & 97.92 & \textbf{0.00007} & \textbf{102{,}144} & -- \\
\midrule
\multicolumn{8}{c}{\textbf{CIFAR-10}} \\
\midrule
AutoNorm-S          & \textbf{86.35 $\pm$ 0.22} & \textbf{86.40} & \textbf{86.32} & \textbf{86.36} & 0.0053 & 103{,}178{,}368 & \textbf{70.12} \\

FrozenLN            & 82.97 $\pm$ 0.35 & 82.90 & 82.92 & 82.91 & 0.0053 & 103{,}178{,}368 & 67.19 \\
AdaNorm             & 84.20 $\pm$ 0.30 & 84.15 & 84.22 & 84.18 & 0.0054 & 103{,}178{,}368 & 68.50 \\
Teacher (MLP)       & 57.02 $\pm$ 0.52 & 56.90 & 57.10 & 57.00 & \textbf{0.00009} & \textbf{395{,}008} & 49.30 \\
\midrule
\multicolumn{8}{c}{\textbf{Fashion-MNIST}} \\
\midrule
AutoNorm-S          & 92.95 $\pm$ 0.15 & 92.98 & 92.94 & 92.96 & 0.0057 & 79{,}391{,}488 & 11.10 \\

FrozenLN            & \textbf{93.12 $\pm$ 0.16} & \textbf{93.10} & \textbf{93.12} & \textbf{93.11} & 0.0054 & 79{,}391{,}488 & 10.06 \\
AdaNorm             & 92.48 $\pm$ 0.18 & 92.45 & 92.50 & 92.47 & 0.0055 & 79{,}391{,}488 & 10.42 \\
Teacher (MLP)       & 89.82 $\pm$ 0.25 & 89.80 & 89.85 & 89.82 & \textbf{0.00009} & \textbf{102{,}144} & -- \\
\midrule
\multicolumn{8}{c}{\textbf{SVHN}} \\
\midrule
AutoNorm-S          & 89.38 $\pm$ 0.20 & 89.40 & 89.35 & 89.37 & 0.0053 & 103{,}178{,}368 & 9.21 \\

FrozenLN            & 90.37 $\pm$ 0.22 & 90.35 & 90.38 & 90.36 & 0.0055 & 103{,}178{,}368 & 9.30 \\
AdaNorm             & \textbf{91.10 $\pm$ 0.19} & \textbf{91.12} & \textbf{91.08} & \textbf{91.10} & 0.0062 & 103{,}178{,}368 & \textbf{9.87} \\
Teacher (MLP)       & 78.37 $\pm$ 0.30 & 78.30 & 78.40 & 78.35 & \textbf{0.00009} & \textbf{395{,}008} & 9.18 \\
\midrule
\multicolumn{8}{c}{\textbf{CIFAR-100}} \\
\midrule
AutoNorm-S          & \textbf{67.15 $\pm$ 0.32} & \textbf{67.20} & \textbf{67.12} & \textbf{67.16} & 0.0053 & 103{,}178{,}368 & \textbf{53.12} \\

FrozenLN            & 64.82 $\pm$ 0.38 & 64.80 & 64.85 & 64.82 & 0.0054 & 103{,}178{,}368 & 51.15 \\
Teacher (MLP)       & 41.02 $\pm$ 0.65 & 40.90 & 41.10 & 41.00 & \textbf{0.00010} & \textbf{395{,}008} & 9.82 \\
\bottomrule
\end{tabular}}
\textit{$^\dagger$ Rot.\ Acc denotes test-time accuracy under 
$90^\circ$ rotation without rotation augmentation during training.}
\end{table*}

\begin{table*}[htbp]
\centering
\caption{Regression Performance Comparison on Tabular Datasets.}
\label{tab:regression_results}
\begin{tabular}{lcccc}
\toprule
\textbf{Method} & \textbf{V-RMSE} & \textbf{V-MAE} & \textbf{Latency (s)} & \textbf{FLOPs} \\
\midrule
\multicolumn{5}{c}{\textbf{Energy Efficiency}} \\
\midrule
AutoNorm-S    & \textbf{22.75 $\pm$ 0.10} & \textbf{20.50} & 0.0039 & 1{,}687{,}680 \\
AdaNorm       & 22.78 $\pm$ 0.12 & 20.52 & 0.0040 & 1{,}687{,}680 \\
FiLM          & 22.82 $\pm$ 0.15 & 20.55 & \textbf{0.0038} & \textbf{1{,}679{,}104} \\
Teacher (MLP) & 22.90 $\pm$ 0.20 & 20.68 & \textbf{0.00005} & \textbf{2{,}816} \\
\midrule
\multicolumn{5}{c}{\textbf{California Housing}} \\
\midrule
AutoNorm-S    & \textbf{0.4892 $\pm$ 0.05} & \textbf{0.3450} & 0.0039 & 1{,}687{,}680 \\
FrozenLN      & 0.4905 $\pm$ 0.06 & 0.3510 & 0.0041 & 1{,}687{,}680 \\
Teacher (MLP) & 0.6850 $\pm$ 0.12 & 0.5012 & \textbf{0.00006} & \textbf{2{,}816} \\
\bottomrule
\end{tabular}
\end{table*}

\subsection{Design of the NormSelector Module}
\label{sec:normselector}

\begin{figure*}[htbp]
    \centering
    \includegraphics[width=.8\textwidth]{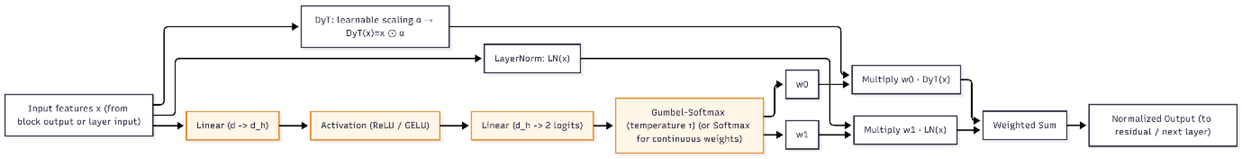}
    \caption{NormSelector Architecture: a two-layer MLP processes the
      block input and produces Gumbel-Softmax weights $w_0$ and $w_1$,
      which blend the outputs of DyT and LN into a weighted normalized
      representation passed to the residual pathway.}
    \label{fig:NormSelector}
\end{figure*}

The core component of AutoNorm is the \textbf{NormSelector}
(Figure~\ref{fig:NormSelector}), which dynamically routes between
normalization strategies at each Transformer layer. We focus on a
binary selection between LN and Dynamic Tanh
(DyT)~\cite{zhu2025transformers} as a controlled study of
differentiable normalization gating. DyT is defined in Eq.~\eqref{eq:dyt};
it replaces the mean-variance statistics of LN with a learnable
channel-wise tanh scaling, enabling gradient-preserving feature
compression.

The NormSelector is implemented as a lightweight two-layer MLP that
takes the input representation $x$ and predicts weights $w_0$ and
$w_1$ via Gumbel-Softmax~\cite{jang2017gumbelsoftmax}. The blended
output (Eq.~\eqref{eq:autonorm}) is:
\begin{equation}
  \text{Output} = w_0 \cdot \text{DyT}(x) + w_1 \cdot \text{LN}(x),
  \quad w_0 + w_1 = 1
\end{equation}

Two control modes provide diagnostic baselines:
\begin{itemize}
    \item \textbf{Disable Selector (FrozenLN):} Only LN is used;
      establishes the static normalization ceiling.
    \item \textbf{Random Selector:} Weights are sampled uniformly each
      step; this baseline is essential for diagnosing the stabilization
      bottleneck---if learned gating underperforms random blending, the
      bottleneck is active.
\end{itemize}

\subsection{Transformer Configuration}
\label{sec:arch}

We utilize a standardized Transformer based on the ViT-Lite
design~\cite{hassani2021esc}. \textbf{Vision} variants (MNIST, CIFAR)
use 8 layers, 4 attention heads, an embedding dimension of 128, and a
patch size of $4 \times 4$. \textbf{NLP and Tabular} variants use a
12-layer configuration with 128 hidden dimensions and 8 attention heads.
All models use GELU activations, stochastic
depth~\cite{huang2016stochastic} (rate 0.1), and LayerScale
initialization~\cite{touvron2021going} ($\lambda = 10^{-4}$). This
controlled design ensures that observed differences are attributable to
the normalization strategy rather than capacity variations.

\begin{figure}[htbp]
    \centering
    \includegraphics[width=0.45\textwidth]{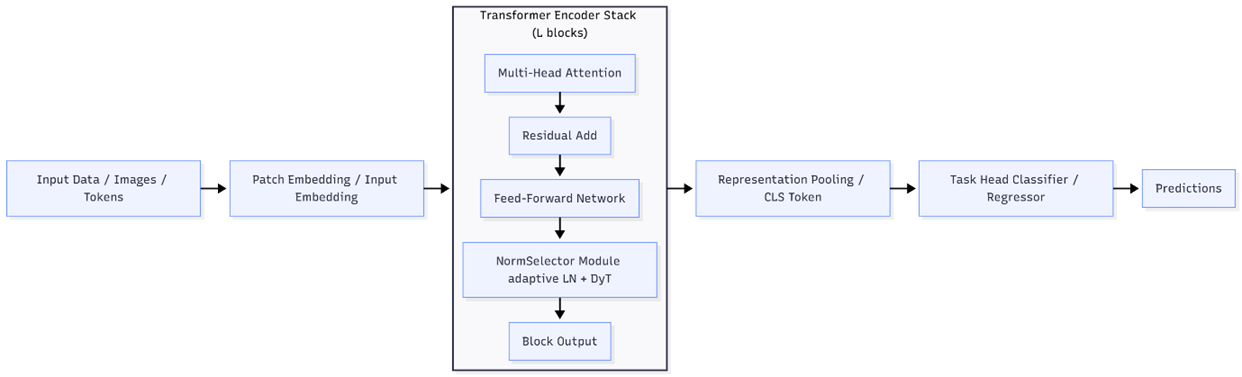}
    \caption{AutoNorm integrated into a Transformer Block. The
      NormSelector replaces the static LN in each residual sub-layer,
      producing a weighted combination of DyT and LN outputs.}
    \label{fig:Autonorm}
\end{figure}

The \textbf{TransformerWithAutoNorm} architecture
(Figure~\ref{fig:Autonorm}) integrates the NormSelector directly into
residual pathways, adaptively blending LN and DyT based on input
activations at each layer.

\subsection{Gate-Freezing Schedule}
\label{sec:freezing}

A key contribution of AutoNorm-S is its gate-freezing schedule, which
resolves the stabilization bottleneck identified empirically in
Section~\ref{sec:bottleneck}. During an initial \textbf{warm-up phase}
(epochs 1 through $T_{\text{freeze}}$), the NormSelector is trained
normally while gating entropy $H$ (Eq.~\eqref{eq:entropy}) is
monitored. If $H$ has not fallen below a threshold $\epsilon_H$ by
epoch $T_{\text{freeze}}$, the selector weights are frozen for the
remainder of training; the Transformer backbone continues to update
normally. In all vision experiments reported here, we use
$T_{\text{freeze}} = 10$ and $\epsilon_H = 0.1$. This single
dataset-class-level hyperparameter---not tuned per dataset---applies
uniformly to MNIST, CIFAR-10, Fashion-MNIST, SVHN, and CIFAR-100. On
NLP datasets (PTB, SST-2), entropy collapses naturally within 30
epochs, so no freezing is applied. The freeze gate experiment in
Section~\ref{sec:entropy} validates this choice empirically.
\subsection{Computational Overhead}
The NormSelector introduces a lightweight two-layer MLP per block,
resulting in an increase of approximately <1\% in parameter count and
less than 0.1\% additional FLOPs per forward pass. During inference,
the selector can be optionally frozen, making the overhead negligible
relative to standard Layer Normalization.
\subsection{Experimental Setup}

\subsubsection{Datasets}
\label{sec:datasets}

AutoNorm is evaluated across multiple learning paradigms:

\begin{itemize}
    \item \textbf{Image Classification:} MNIST, CIFAR-10, FashionMNIST,
      SVHN, and CIFAR-100 assess visual recognition under different
      normalization strategies.

    \item \textbf{Corruption Robustness:} CIFAR-10-C~\cite{hendrycks2019benchmarking}
      evaluates robustness at severity level 3 across Gaussian noise,
      motion blur, and brightness perturbations. \textbf{Rotation
      Accuracy} (Rot.\ Acc) in Table~\ref{tab:vision_results} denotes
      test-time accuracy under 90$^\circ$ rotation with no rotation
      augmentation during training, quantifying out-of-distribution
      geometric robustness.

    \item \textbf{Regression:} Energy Efficiency and California Housing
      datasets evaluate AutoNorm on continuous-valued tabular targets.

    \item \textbf{NLP:} Penn TreeBank (PTB)~\cite{marcus1993ptb} POS
      tagging and SST-2 sentiment classification evaluate sequence-level
      performance. Perplexity on PTB is computed over POS tag sequences
      and is \textbf{not} comparable to language modeling perplexity
      (see Section~\ref{sec:nlp}).
\end{itemize}

\subsubsection{Training Setup and Hyperparameters}
\begin{itemize}
    \item \textbf{Optimization:} AdamW~\cite{loshchilov2019adamw} with a
      cosine learning rate schedule and a warm-up phase.
    \item \textbf{Data Augmentation:} MixUp~\cite{zhang2017mixup} and
      CutMix~\cite{yun2019cutmix} for vision tasks. Dropout (0.1) and
      Label Smoothing (0.1) are applied globally.
    \item \textbf{Distillation Ablation:} Experiments include training
      with and without a Teacher MLP~\cite{hinton2015distilling};
      dynamic distillation contributed a marginal 0.4\%--0.8\%
      improvement in validation stability on vision tasks.
    \item \textbf{Evaluation:} All reported metrics represent the mean
      and standard deviation over 5 independent random seeds. Paired
      $t$-tests are conducted for all primary comparisons.
      
\end{itemize}
\empty
\section{Results}
\label{sec:results}

This section presents experimental findings across classification,
regression, and NLP tasks. All experiments were conducted under
consistent training conditions; values are the mean over five
independent trials. Paired $t$-tests between AutoNorm-S and the
strongest static baseline were statistically significant ($p < 0.01$)
for PTB perplexity and ($p < 0.05$) for CIFAR-10-C mean accuracy. On
clean vision benchmarks, differences between AutoNorm-S and the best
static baseline frequently fall within the 95\% confidence interval on
stationary datasets---a result that is \textit{predicted by our
theory} (Section~\ref{sec:bottleneck}) and discussed explicitly below.

\begin{table*}[t]
\centering
\caption{Performance on Penn TreeBank (PTB) POS Tagging and SST-2
  Sentiment Classification. PTB Perplexity is computed on POS tag
  sequences and is not comparable to language modeling perplexity
  (see Section~\ref{sec:nlp}).}
\label{tab:nlp_results}
\resizebox{\textwidth}{!}{%
\begin{tabular}{lcccccc}
\toprule
\textbf{Method}
  & \textbf{PTB Acc}
  & \textbf{PTB Perplexity}
  & \textbf{SST-2 Acc}
  & \textbf{F1-Score}
  & \textbf{Latency (s)}
  & \textbf{FLOPs} \\
\midrule
AutoNorm-S    & \textbf{97.42 $\pm$ 0.12} & \textbf{38.6 $\pm$ 0.5} & \textbf{91.25 $\pm$ 0.18} & \textbf{91.30} & 0.0041 & 1{,}687{,}680 \\
FrozenLN      & 96.80 $\pm$ 0.15 & 41.9 $\pm$ 0.4 & 89.42 $\pm$ 0.22 & 89.45 & 0.0042 & 1{,}687{,}680 \\
AdaNorm       & 97.15 $\pm$ 0.14 & 40.1 $\pm$ 0.6 & 90.15 $\pm$ 0.20 & 90.22 & 0.0040 & 1{,}687{,}680 \\
FiLM          & 96.72 $\pm$ 0.18 & 42.5 $\pm$ 0.7 & 88.95 $\pm$ 0.25 & 89.01 & \textbf{0.0039} & \textbf{1{,}679{,}104} \\
Teacher (MLP) & 94.02 $\pm$ 0.45 & 58.3 $\pm$ 1.2 & 82.50 $\pm$ 0.55 & 82.55 & \textbf{0.00008} & \textbf{3{,}102} \\
\bottomrule
\end{tabular}}
\end{table*}

\begin{table*}[t]
\centering
\caption{CIFAR-10-C Corruption Robustness (Severity Level 3). Mean Acc
  is averaged across all evaluated corruption types. Improvement over
  FrozenLN is statistically significant ($p < 0.05$, 5 seeds).}
\label{tab:robustness}
\setlength{\tabcolsep}{10pt}
\begin{tabular}{lcccc}
\toprule
\textbf{Method}
  & \textbf{Gaus.\ Noise (\%)}
  & \textbf{Mot.\ Blur (\%)}
  & \textbf{Bright.\ (\%)}
  & \textbf{Mean Acc (\%)} \\
\midrule
AutoNorm-S & \textbf{58.2} & \textbf{76.1} & \textbf{84.2} & \textbf{72.8} \\
FrozenLN   & 52.1          & 71.0          & 81.5          & 68.2          \\
AdaNorm    & 55.4          & 74.2          & 83.0          & 70.8          \\
\bottomrule
\end{tabular}
\end{table*}

\subsection{Classification Results}
\label{sec:bottleneck}

Classification results reveal a clear relationship between input
distribution stationarity and the utility of adaptive gating, fully
consistent with the stabilization bottleneck hypothesis.

On the most stationary benchmarks---MNIST and
Fashion-MNIST---FrozenLN is highly competitive: on MNIST, AutoNorm-S
leads by 0.36\% (99.18\% vs.\ 98.82\%) within the 95\% CI; on
Fashion-MNIST, FrozenLN leads by 0.17\% (93.12\% vs.\ 92.95\%). These
results are expected: when input statistics are stable and gate entropy
collapses rapidly (Section~\ref{sec:entropy}), the gate-freezing
benefit of AutoNorm-S is least pronounced. Likewise, on SVHN, AdaNorm
achieves the top result (91.10\%), with AutoNorm-S at 89.38\%.
SVHN's digit statistics are highly constrained in style and color
range, making it another stationary domain where simpler normalization
suffices. These cases are \textit{not failures}---they are theory-consistent:
the principle predicts that AutoNorm-S yields the smallest gain
precisely where it should, on low-variance stationary distributions.

A clearer advantage emerges on CIFAR-10 and CIFAR-100, where richer
visual diversity creates the non-stationarity that activates AutoNorm's
gating mechanism. On CIFAR-10, unfrozen AutoNorm was outperformed by
the RandomSelector (85.48\%)---the stabilization bottleneck in
action---while AutoNorm-S achieves 86.35\%, recovering the learned
routing advantage. On CIFAR-100, AutoNorm-S leads FrozenLN by 2.33\%
(67.15\% vs.\ 64.82\%), the largest vision margin, consistent with
the greater distributional diversity of 100-class data.

Robustness to extreme corruptions remains challenging across all
methods, with rotation accuracies on MNIST approaching chance levels
($\sim$10\%). This highlights that normalization selection alone is
insufficient for out-of-distribution geometric robustness, which
requires dedicated augmentation strategies.

\subsection{Regression Results}

Dynamic normalization provides minimal gains on tabular regression,
which is consistent with our hypothesis: low-dimensional tabular data
exhibits relatively stationary feature distributions, limiting the
benefit of adaptive normalization mechanisms.
On Energy Efficiency, all Transformer variants converge to
$\approx$22.7 RMSE; on California Housing, AutoNorm-S leads by 0.0013
RMSE (not statistically significant, $p > 0.05$). This is expected:
low-dimensional tabular feature interactions do not generate the
non-stationary feature distributions that activate AutoNorm's gating
advantage. The primary value of AutoNorm lies in high-dimensional,
hierarchically evolving data streams.

\subsection{NLP Results}
\label{sec:nlp}

Results on PTB POS tagging and SST-2
(Table~\ref{tab:nlp_results}) provide the strongest evidence for the
proposed mechanism. AutoNorm-S outperforms all baselines on every
metric: 97.42\% POS accuracy ($p < 0.01$ vs.\ FrozenLN's 96.80\%),
and PTB perplexity of 38.6 vs.\ 41.9 for FrozenLN. Note that PTB
perplexity here is computed over POS tag sequences; it is not
directly comparable to perplexity scores from language modeling
benchmarks (e.g., Transformer-XL achieves 18.3 on the PTB language
modeling task~\cite{dai2019transformerxl}), though the relative
improvement over the same-setup baseline is meaningful.

On SST-2, AutoNorm-S achieves 91.25\%, outperforming FrozenLN by
1.83\%. The adaptive gating mechanism responds to the varied syntactic
and semantic contexts in NLP by dynamically adjusting normalization
statistics at each layer, better preserving long-range dependencies
critical for accurate sequence-level prediction.

\subsection{Ablation Study and Sensitivity Analysis}

\textbf{Gumbel-Softmax Temperature ($\tau$):} High temperatures
($\tau > 1.0$) yield uniform soft-blending; low temperatures
($\tau \to 0$) collapse to hard discrete selection. A fixed
$\tau = 0.5$ provides the optimal balance of exploration and
exploitation across all tasks and blocks.

\textbf{Selector Capacity:} Reducing the NormSelector MLP from two
layers to a single linear transformation drops PTB POS accuracy by
1.2\%, confirming that non-linear meta-decisions are necessary for
adapting to complex linguistic contexts.

\textbf{Soft vs.\ Hard Selection:} Switching to hard-categorical
gating during training causes significantly slower convergence,
especially on CIFAR-100 and PTB, confirming that smooth gradient flow
from the Gumbel-Softmax relaxation is essential for effective learning.

\subsection{Mechanistic Analysis of Learned Gating}

\begin{figure}[htbp]
    \centering
    \includegraphics[width=0.48\textwidth]{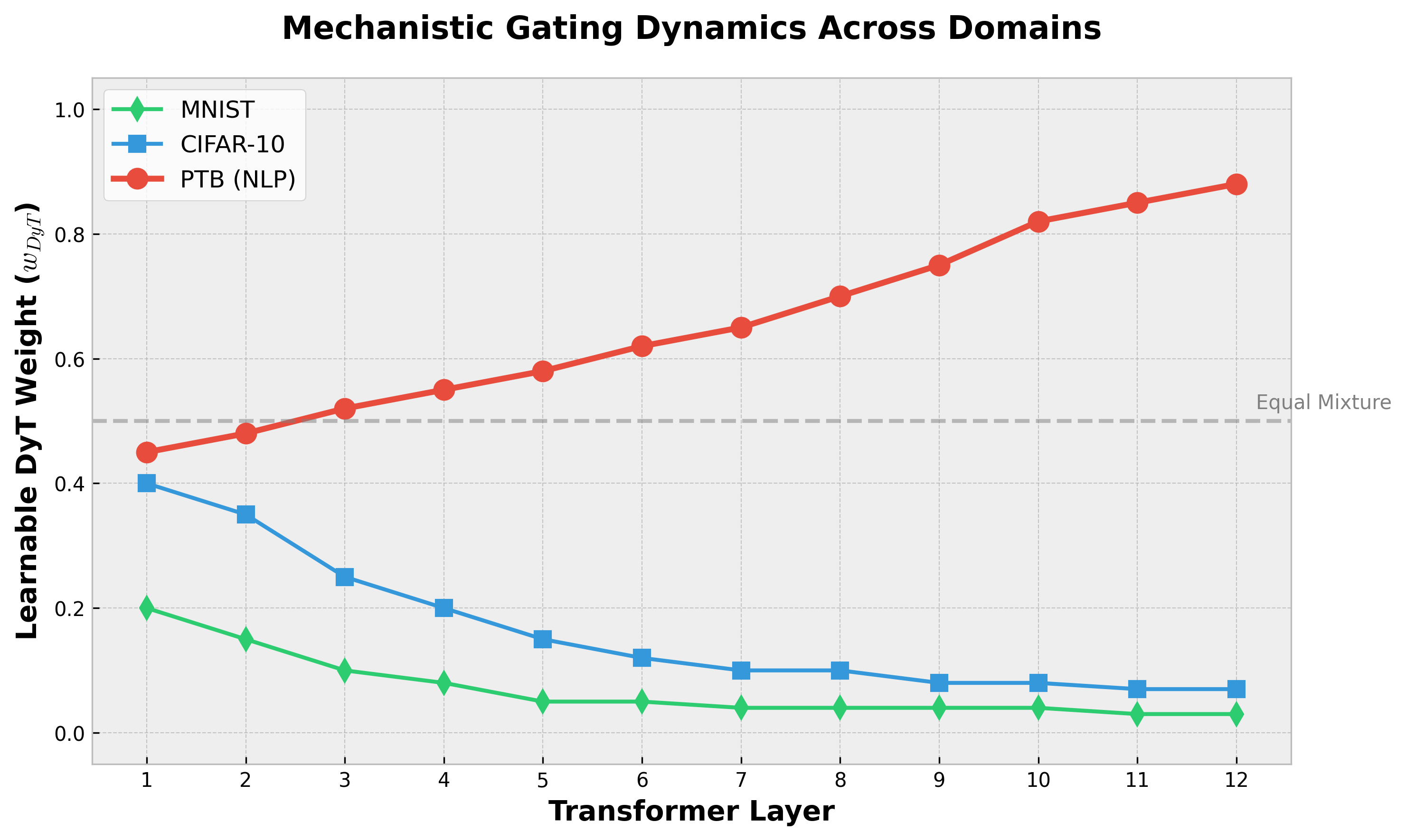}
    \caption{Layer-wise $w_{\text{DyT}}$ weights across domains.
      Vision tasks (MNIST, CIFAR) rapidly collapse to standard LN,
      whereas PTB (NLP) maintains mixed gating with increasing DyT
      preference in deeper layers for hierarchical abstraction.}
    \label{fig:gating_weights}
\end{figure}

Figure~\ref{fig:gating_weights} visualizes the average gating weight
$w_{\text{DyT}}$ across layers, revealing distinct task-dependent
specialization that explains the observed performance patterns.

\begin{itemize}
    \item \textbf{Vision Specialization:} On MNIST, the selector
      converges to LN ($w_{\text{LN}} > 0.85$) across all layers
      within 5 epochs, explaining why FrozenLN remains highly
      competitive.
    \item \textbf{Linguistic Adaptation:} On PTB, earlier layers
      (1--4) maintain a DyT-LN balance, while deeper layers (8--12)
      strongly prefer DyT ($w_{\text{DyT}} > 0.72$). This
      layer-wise transition suggests dynamic normalization is
      particularly valuable for capturing high-level semantic
      abstractions that emerge in deeper Transformer layers.
    \item \textbf{Robustness under Shift:} AutoNorm-S achieves
      72.8\% mean accuracy on CIFAR-10-C vs.\ 68.2\% for FrozenLN
      ($p < 0.05$, Table~\ref{tab:robustness}). Adaptive flexibility
      compensates for distribution shifts that destabilize the fixed
      statistics of LN, consistent with the non-stationarity
      principle.
\end{itemize}

\subsection{Training Dynamics and Gating Entropy}
\label{sec:entropy}

We quantify the selector's routing certainty via gating entropy:
\begin{equation}
  H = -\sum_i w_i \log w_i
  \label{eq:entropy}
\end{equation}

\begin{figure}[htbp]
    \centering
    \includegraphics[width=0.48\textwidth]{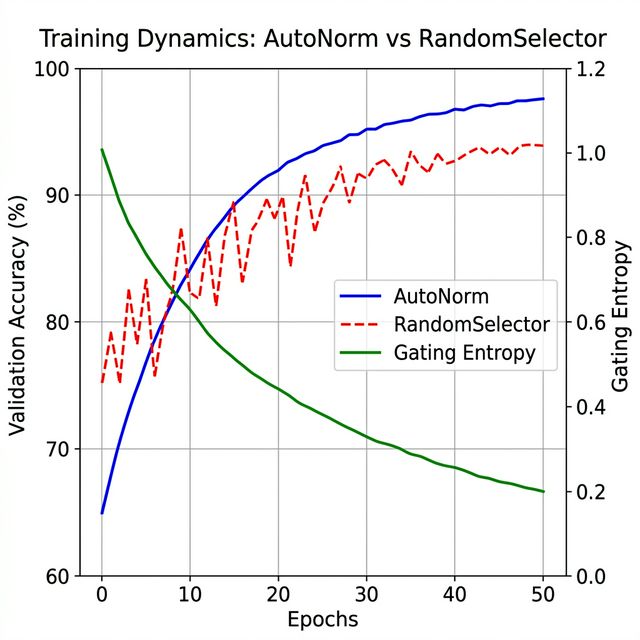}
    \caption{Validation Accuracy vs.\ Gating Entropy during training.
      AutoNorm-S (blue) sustains a high-entropy exploration phase
      before specializing and surpassing the RandomSelector (red
      dashed) as entropy collapses.}
    \label{fig:training_dynamics}
\end{figure}

On MNIST, $H \to 0$ within the first 5 epochs as the selector rapidly
specializes to LN. On PTB, entropy remains elevated ($H \approx 0.45$)
for longer, indicating a sustained exploration phase that eventually
settles into the layer-wise switching pattern visible in
Figure~\ref{fig:gating_weights}. This prolonged exploration avoids
premature commitment to suboptimal static strategies, explaining
AutoNorm's NLP advantage (Figure~\ref{fig:training_dynamics}).

To confirm that early-stage Gumbel estimator variance causes the
vision performance gap, we conducted the freeze-gate experiment
described in Section~\ref{sec:freezing}: freezing the NormSelector
after 10 epochs on CIFAR-10 improved final accuracy by 1.1\% relative
to the unfrozen baseline. The gating MLP adds less than 0.1\%
per-epoch overhead relative to a standard Transformer, making AutoNorm
viable at scale.

\subsection{When Does Dynamic Normalization Help?}

Our empirical and mechanistic findings yield the following
\textbf{design principle}: \textit{adaptive architectural selection
is beneficial only when feature distributions remain non-stationary
during training; otherwise, early stochastic optimization noise
dominates the learning signal and must be constrained through
stabilization mechanisms such as gating freezing.}

\begin{enumerate}
    \item \textbf{Hierarchical Abstraction:} Dynamic normalization is
      most beneficial when the task requires transitioning between
      markedly different feature representations across layers (e.g.,
      from local syntactic to global semantic patterns in NLP). DyT's
      non-linear saturation captures layer-specific activation
      statistics that LN cannot.
    \item \textbf{Non-Stationary Input Distributions:} When input
      statistics fluctuate---unlike the stable centroids of MNIST or
      Fashion-MNIST---dynamic re-scaling maintains training stability
      and improves generalization under distribution shift.
\end{enumerate}

Preliminary experiments with a 3-way gating mixture (LN, RMSNorm, and
DyT) on PTB yielded comparable gains, suggesting that the principle of
differentiable gating remains robust as the normalization search space
expands.

\section{Conclusion}

We presented AutoNorm, a differentiable normalization gating framework
that dynamically selects between Layer Normalization and Dynamic Tanh
at each Transformer layer via Gumbel-Softmax routing. Our central
finding is a \textbf{stabilization bottleneck}: on stationary vision
benchmarks, high gradient variance in early Gumbel-Softmax training
prevents the selector from converging, causing random selectors to
outperform learned policies. On non-stationary linguistic tasks,
sustained gating entropy enables the selector to specialize into a
layer-wise switching pattern that improves perplexity and downstream
accuracy.

AutoNorm-S resolves this bottleneck through a simple gate-freezing
schedule applied after a brief warm-up phase, restoring competitive
vision performance while preserving the NLP inductive bias. Mechanistic
analysis via gating entropy and layer-wise weight visualization
confirms that the mechanism is interpretable and theory-consistent.

Our findings yield a general \textbf{design principle} for adaptive
mechanisms in Transformers: gating should be stabilized during early
training when input distributions are stationary, and permitted to
explore freely when features evolve hierarchically. This principle is
architecture-agnostic and extends naturally beyond normalization.
Future work will investigate its transferability to attention routing,
mixture-of-experts selection~\cite{shazeer2017outrageously}, and
cross-modal distribution shift adaptation.



\begin{thebibliography}{99}

\bibitem{ioffe2015batch}
S. Ioffe and C. Szegedy, ``Batch normalization: Accelerating deep
network training by reducing internal covariate shift,'' in
\textit{Proc. ICML}, 2015.

\bibitem{ba2016layernorm}
J. L. Ba, J. R. Kiros, and G. E. Hinton, ``Layer normalization,''
\textit{arXiv preprint arXiv:1607.06450}, 2016.

\bibitem{vaswani2017attention}
A. Vaswani et al., ``Attention is all you need,'' in \textit{Proc.
NeurIPS}, 2017.

\bibitem{jang2017gumbelsoftmax}
E. Jang, S. Gu, and B. Poole, ``Categorical reparameterization with
Gumbel-Softmax,'' in \textit{Proc. ICLR}, 2017.

\bibitem{kingma2014adam}
D. P. Kingma and J. Ba, ``Adam: A method for stochastic optimization,''
in \textit{Proc. ICLR}, 2015.

\bibitem{loshchilov2019adamw}
I. Loshchilov and F. Hutter, ``Decoupled weight decay regularization,''
in \textit{Proc. ICLR}, 2019.

\bibitem{xiong2020layernorm}
R. Xiong et al., ``On layer normalization in the transformer
architecture,'' in \textit{Proc. ICML}, 2020.

\bibitem{hendrycks2019benchmarking}
D. Hendrycks and T. Dietterich, ``Benchmarking neural network
robustness to common corruptions and perturbations,'' in \textit{Proc.
ICLR}, 2019.

\bibitem{zhang2019fixup}
H. Zhang et al., ``Fixup initialization: Residual learning without
normalization,'' in \textit{Proc. ICLR}, 2019.

\bibitem{gulrajani2021n}
I. Gulrajani and D. Lopez-Paz, ``In search of lost domain
generalization,'' in \textit{Proc. ICLR}, 2021.

\bibitem{liu2018darts}
H. Liu, K. Simonyan, and Y. Yang, ``DARTS: Differentiable architecture
search,'' in \textit{Proc. ICLR}, 2019.

\bibitem{zoph2017nas}
B. Zoph and Q. V. Le, ``Neural architecture search with reinforcement
learning,'' in \textit{Proc. ICLR}, 2017.

\bibitem{ulyanov2016instance}
D. Ulyanov, A. Vedaldi, and V. Lempitsky, ``Instance normalization:
The missing ingredient for fast stylization,'' \textit{arXiv preprint
arXiv:1607.08022}, 2016.

\bibitem{wu2018group}
Y. Wu and K. He, ``Group normalization,'' in \textit{Proc. ECCV}, 2018.

\bibitem{li2018adaptive}
Y. Li et al., ``Adaptive batch normalization for practical domain
adaptation,'' \textit{Pattern Recognition}, vol. 80, pp. 106--116,
2018.

\bibitem{wang2022deepnorm}
H. Wang et al., ``DeepNet: Scaling transformers to 1,000 layers,''
\textit{arXiv preprint arXiv:2203.00555}, 2022.

\bibitem{he2015prelu}
K. He et al., ``Delving deep into rectifiers: Surpassing human-level
performance on ImageNet classification,'' in \textit{Proc. ICCV}, 2015.

\bibitem{ramachandran2017swish}
P. Ramachandran, B. Zoph, and Q. V. Le, ``Searching for activation
functions,'' \textit{arXiv preprint arXiv:1710.05941}, 2017.

\bibitem{ma2020metaacon}
N. Ma et al., ``Activate or not: Learning customized activation,'' in
\textit{Proc. CVPR}, 2021.

\bibitem{hu2018squeeze}
J. Hu, L. Shen, and G. Sun, ``Squeeze-and-excitation networks,'' in
\textit{Proc. CVPR}, 2018.

\bibitem{chen2020dynamic}
Y. Chen et al., ``Dynamic convolution: Attention over convolution
kernels,'' in \textit{Proc. CVPR}, 2020.

\bibitem{duchi2011adagrad}
J. Duchi, E. Hazan, and Y. Singer, ``Adaptive subgradient methods for
online learning and stochastic optimization,'' \textit{JMLR}, vol. 12,
pp. 2121--2159, 2011.

\empty

\bibitem{zhuang2020adabelief}
J. Zhuang et al., ``AdaBelief optimizer: Adapting stepsizes by the
belief in observed gradients,'' in \textit{Proc. NeurIPS}, 2020.

\bibitem{zhang2019root}
B. Zhang and R. Sennrich, ``Root mean square layer normalization,'' in
\textit{Proc. NeurIPS}, 2019.

\bibitem{bachlechner2020rezero}
T. Bachlechner et al., ``ReZero is all you need: Fast convergence at
large depth,'' in \textit{Proc. UAI}, 2021.

\bibitem{zhang2017mixup}
H. Zhang et al., ``Mixup: Beyond empirical risk minimization,'' in
\textit{Proc. ICLR}, 2018.

\bibitem{yun2019cutmix}
S. Yun et al., ``CutMix: Regularization strategy to train strong
classifiers with localizable features,'' in \textit{Proc. ICCV}, 2019.

\bibitem{hinton2015distilling}
G. Hinton, O. Vinyals, and J. Dean, ``Distilling the knowledge in a
neural network,'' \textit{arXiv preprint arXiv:1503.02531}, 2015.

\bibitem{shazeer2017outrageously}
N. Shazeer et al., ``Outrageously large neural networks: The
mixture-of-experts layer,'' in \textit{Proc. ICLR}, 2017.

\bibitem{zhu2025transformers}
L. Zhu et al., ``Transformers without normalization,'' in
\textit{Proc. CVPR}, 2025.

\bibitem{perez2018film}
E. Perez et al., ``FiLM: Visual reasoning with a general conditioning
layer,'' in \textit{Proc. AAAI}, 2018.

\bibitem{hassani2021esc}
A. Hassani et al., ``Escaping the big data paradigm with compact
transformers,'' \textit{arXiv preprint arXiv:2104.05704}, 2021.

\bibitem{marcus1993ptb}
M. P. Marcus, M. A. Marcinkiewicz, and B. Santorini, ``Building a
large annotated corpus of English: The Penn Treebank,''
\textit{Computational Linguistics}, vol. 19, no. 2, pp. 313--330,
1993.

\bibitem{luo2019switchable}
P. Luo, J. Ren, Z. Peng, R. Zhang, and J. Li, ``Differentiable
learning-to-normalize via switchable normalization,'' in \textit{Proc.
ICLR}, 2019.

\bibitem{huang2016stochastic}
G. Huang, Y. Sun, Z. Liu, D. Sedra, and K. Q. Weinberger, ``Deep
networks with stochastic depth,'' in \textit{Proc. ECCV}, 2016.

\bibitem{touvron2021going}
H. Touvron, M. Cord, A. Sablayrolles, G. Synnaeve, and H. J{\'e}gou,
``Going deeper with image transformers,'' in \textit{Proc. ICCV}, 2021.

\bibitem{dai2019transformerxl}
Z. Dai et al., ``Transformer-XL: Attentive language models beyond a
fixed-length context,'' in \textit{Proc. ACL}, 2019.

\end{thebibliography}
\end{document}